\documentclass{article}
\usepackage{authblk}

\usepackage{times}
\usepackage{graphicx} 

\usepackage{natbib}

\usepackage{algorithm}
\usepackage{algorithmic}

\usepackage[algo2e,ruled,vlined,linesnumbered]{algorithm2e}
\usepackage{tabularx}
\usepackage{epsfig}
\usepackage{amssymb}
\usepackage{amsmath}
\usepackage{amsfonts}

\usepackage{lscape}
\usepackage{subcaption}

\usepackage{hyperref}

\makeatletter
\newcommand{\mathleft}{\@fleqntrue\@mathmargin0pt}
\newcommand{\mathcenter}{\@fleqnfalse}
\makeatother

\begin{document}

\newcommand{\norm}[1]{\left\lVert#1\right\rVert}
\newcommand*{\Scale}[2][4]{\scalebox{#1}{$#2$}}%
\newcommand*{\Resize}[2]{\resizebox{#1}{!}{$#2$}}%

\newcommand{\stream}{stream}
\newcommand{\coreset}{\textsc{Coreset}}

\newcommand{\abs}[1]        {\left| #1\right|}
\providecommand{\norm}[1]{\left\lVert#1\right\rVert}
\providecommand{\inn}[1]{\langle#1\rangle}
\newcommand{\br}[1]{\left\{#1\right\}}
\newcommand{\REAL}{\ensuremath{\mathbb{R}}}
\newcommand{\eps}{\varepsilon}
\newcommand{\xs}{z}
\newcommand{\y}{y}
\newcommand{\nnz}{\mathrm{nnz}}

\newtheorem{theorem}{Theorem}
\newtheorem{problem}{Problem}
\newtheorem{definition}{Definition}
\newtheorem{lemma}{Lemma}
\newtheorem{remark}{Remark}
\newtheorem{claim}{Claim}
\newtheorem{corollary}{Corollary}

\newcommand{\dist}{\mathrm{dist}}
\newcommand{\dimm}{\mathrm{dim}}
\newcommand{\Frank}{\textsc{Frank-Wolf}}
\newcommand{\corealgA}{\textsc{CoresetI}}
\newcommand{\corealgB}{\textsc{CoresetII}}
\newcommand{\corealgC}{\textsc{CoresetPCA}}
\newcommand{\epp}{\eps}
\newcommand{\N}{N}
\renewcommand{\j}{j}
\newcommand{\fix}{\marginpar{FIX}}
\newcommand{\new}{\marginpar{NEW}}
\newcommand{\SVDCoreset}{\textsc{SVD-Coreset}}
\newcommand{\svds}{\textsc{svds}}
\newcommand{\qr}{\textsc{qr}}
\newcommand{\pe}{\psi}
\newcommand{\Four}{{\mathcal{F}}}
\newcommand{\Fourinv}{{\mathcal{F}}^{-1}}
\newcommand{\set}[1]        {\left\{ #1 \right\}}
\newcommand{\mathsc}[1]{\text{\textsc{#1}}}
\newcommand{\rbr}[1]      {\left( #1 \right)}
\newcommand{\floor}[1]      {\left\lfloor #1 \right\rfloor}
\newcommand{\ceil}[1]       {\left\lceil #1 \right\rceil}
\newcommand{\ang}[1]        {\left\langle #1 \right\rangle}
\newcommand{\sbr}[1]	    {\left[#1\right]}
\newcommand{\cbr}[1]		 {\left\{#1\right\}}
\newcommand{\cost}{\mathrm{cost}}
\def\GR{\textbf{GR:}}
\def\d{\delta}
\newcommand{\T}{{}^\mathsf{T}}
\def\ds{\displaystyle}
\def\e{{\epsilon}}
\def\eb{\bar{\eta}}
\def\enorm#1{\|#1\|_2}
\def\Fp{F^\prime}
\def\fishpack{{FISHPACK}}
\def\fortran{{FORTRAN}}
\def\gmres{{GMRES}}
\def\gmresm{{\rm GMRES($m$)}}
\def\Kc{{\cal K}}
\def\wb{{\bar w}}
\def\zb{{\bar z}}
\newcommand{\RR}{\mathbb{R}}
\newcommand{\RD}{\mathbb{R}^d}
\newcommand{\NN}{\mathbb{N}}
\newcommand{\cdc}{,\,\ldots,\,}
\newcommand{\msize}{j+j/\eps} 
\newcommand{\mycomment}[1]{{\footnotesize\textit{//#1}}}
\newcommand{\absv}[1]{\left| #1\right|}

\newcommand{\var}{\mathrm{var}}

\setlength{\abovedisplayskip}{3pt}
\setlength{\belowdisplayskip}{3pt}





%
%

\author[1]{Dan Feldman}
\author[2]{Sedat Ozer}
\author[2]{Daniela Rus}
\affil[1]{Robotics \& Big Data Lab\\ University of Haifa}
\affil[2]{CSAIL, MIT}
\setcounter{Maxaffil}{0}
\renewcommand\Affilfont{\itshape\small}

\title{Coresets for Vector Summarization with Applications to Network Graphs}
\date{}
\maketitle

\begin{abstract}
We provide a deterministic data summarization algorithm that approximates the mean $\bar{p}=\frac{1}{n}\sum_{p\in P} p$ of a set $P$ of $n$ vectors in $\REAL^d$, by a weighted mean $\tilde{p}$ of a \emph{subset} of $O(1/\eps)$ vectors, i.e., independent of both $n$ and $d$. We prove that the squared Euclidean distance between $\bar{p}$ and $\tilde{p}$ is at most $\eps$ multiplied by the variance of $P$. We use this algorithm to maintain an approximated sum of vectors from an unbounded stream, using memory that is independent of $d$, and logarithmic in the $n$ vectors seen so far. Our main application is to extract and represent in a compact way friend groups and activity summaries of users from underlying data exchanges. For example, in the case of mobile networks, we can use GPS traces to identify meetings; in the case of social networks, we can use information exchange to identify friend groups. Our algorithm provably identifies the {\it Heavy Hitter} entries in a proximity (adjacency) matrix. The Heavy Hitters can be used to extract and represent in a compact way friend groups and activity summaries of users from underlying data exchanges. We evaluate the algorithm on several large data sets.
\end{abstract}


\section{Introduction}

The wide-spread use of smart phones, wearable devices, and social media creates a vast space of digital footprints for people, which include location information from GPS traces, phone call history, social media postings, etc. This is an ever-growing wealth of data that can be used to identify social structures and predict activity patterns. We wish to extract the underlying social network of a group of mobile users given data available about them (e.g. GPS traces, phone call history, news articles, etc.) in order to identify and predict their various activities such as meetings, friend groups, gathering places, collective activity patterns, etc. There are several key challenges to achieve these capabilities. First, the data is huge so we need efficient methods for processing and representing the data. Second, the data is multi-modal heterogeneous. This presents challenges in data processing and representation, but also opportunities to extract correlations that may not be visible in a single data source. Third, the data is often noisy.

We propose an approach based on coresets to extract underlying connectivity information while performing data summarization for a given a large data set. We focus our intuition examples and evaluations on social networks because of their intuitive nature and access to data sets, although the method is general and applies to networks of information in general. Our approach works on streaming datasets to represent the data in a compact (sparse) way. Our coreset algorithm gets a stream of vectors and approximates their sum using small memory. Essentially, a coreset $C$ is a significantly smaller portion (a scaled subset) of the original and large set $D$ of vectors. Given $D$ and the algorithm $A$, where running algorithm $A$ on $D$ is intractable due to lack of memory, the task-specific coreset algorithm efficiently reduces the data set $D$ to a coreset $C$ so that running the algorithm $A$ on $C$ requires a low amount of memory and the result is provable approximately the same as running the algorithm on $D$. Coreset captures all the important vectors in $D$ for a given algorithm $A$.  The challenges are computing $C$ fast and proving that $C$ is the right scaled subset, i.e., running the algorithm on $C$ gives approximately the same result as running the algorithm on $D$.

More specifically, the goal of this paper is to suggest a way to maintain a sparse representation of an $n\times d$ matrix, by maintaining a sparse approximation of each of its rows. For example, in a proximity matrix associated with a social network, instead of storing the average proximity to each of $n$ users, we would like to store only the $N\ll n$ largest entries in each row (known as ``Heavy Hitters"), which correspond to the people seen by the user most often. Given an unbounded stream of movements, it is hard to tell which are the people the user met most, without maintaining a counter to each person. For example, consider the special case $N=1$. We are given a stream of pairs $(i,val)$ where $i\in\{1,\cdots,n\}$ is an index of a counter, and $val$ is a real number that represents a score for this counter. Our goal is to identify which counter has the maximum average score till now, and approximate that score. While that is easy to do by maintaining the sum of $n$ scores, our goal is to maintain only a constant set of numbers (memory words). In general, we wish to have a provable approximation to each of the $n$ accumulated scores (row in a matrix), using, say, only $O(1)$ memory. Hence, maintaining an $n\times n$ matrix would take $O(n)$ instead of $O(n^2)$ memory. For millions of users or rows, this means $1$ Gigabytes of memory can be stored in RAM for real-time updates, compared to millions of Gigabytes. Such data reduction will make it practical to keep hundreds of matrices for different types of similarities (sms, phone calls, locations), different types of users, and different time frames (average proximity in each day, year, etc). This paper contributes the following: (1) A compact representation for streaming proximity data for a group of many users; (2) A coreset construction algorithm for maintaining the social network with error guarantees; (3) An evaluation of the algorithm on several data sets.

\textbf{Theoretical Contribution:}  These results are based on an algorithm that computes an $\eps$-coreset $C$ of size $|C|=O(1/\eps)$ for the mean of a given set, and every error parameter $\eps\in(0,1)$ as defined in Section~\ref{sec:statement}. Unlike previous results, this algorithm is deterministic, maintains a weighted subset of the input vectors (which keeps their sparsity), can be applied on a set of vectors whose both cardinality $n$ and dimension $d$ is arbitrarily large or unbounded. Unlike existing randomized sketching algorithms for summing item frequencies ($1$-sparse binary vectors), this coreset can be used to approximate the sum of arbitrary real vectors, including negative entries (for decreasing counters), dense vectors (for fast updates), fractions (weighted counter) and with error that is based on the variance of the vectors (sum of squared distances to their mean) which might be arbitrarily smaller than existing errors: sum/max of squared/non-squared distances to the origin ($\ell_2/\ell_1/\ell_{\infty}$).

\subsection{Solution Overview}
\label{sec:streamed-social-network}
We implemented a system that demonstrates the use and performance of our suggested algorithm. The system constructs a sparse social graph from the GPS locations of real moving smart-phone users and maintains the graph in a streaming fashion as follows.

\textbf{Input stream:} The input to our system is an unbounded stream of (real-time) GPS points, where each point is represented in the vector format of $(time, user ID, longitude, latitude)$. We maintain an approximation of the average proximity of each user to all the other $n-1$ users seen so far, by using space (memory) that is only logarithmic in $n$. The overall memory would then be near-linear in $n$, in contrast to the quadratic $O(n^2)$ memory that is needed to store the exact average proximity vector for each of the $n$ users. We maintain a dynamic array of the $n$ user IDs seen so far and assume, without loss of generality, that the user IDs are distinct and increasing integers from $1$ to $n$. Otherwise we use a hash table from user IDs to such integers. In general, the system is designed to handle any type of streaming records in the format $(stream ID, v)$ where $v$ is a $d$-dimensional vector of reals, to support the other applications. Here, the goal is to maintain a sparse approximation to the sum of the vectors $v$ that were assigned to each stream ID.
\newcommand{\pos}{pos}
\newcommand{\prox}{prox}

\textbf{Proximity matrix:} We also maintain an (non-sparse) array $\pos$ of length $n$ that stores the current location of each of the $n$ users seen so far. That array forms the current $n^2$ pairs of proximities $\prox(u,v)$ between every pair of users and their current locations $u$ and $v$. These pairs correspond to what we call the \emph{proximity matrix} at time $t$, which is a symmetric adjacency $n\times n$ matrix of a social graph, where the edge weight of a pair of users (an entry in this matrix) is their proximity at the current time $t$. We are interested in maintaining the sum of these proximity matrices over time, which, after division by $n$, will give the average proximity between every two users over time. This is the \emph{average proximity matrix}. Since the average proximity matrix and each proximity matrix at a given time require $O(n^2)$ memory, we cannot keep them all in memory. Our goal is to maintain a \emph{sparse approximation} version of each of row in the average proximity matrix which will use only $O(\log n)$ memory. Hence, the required memory by the system will be $O(n\log n)$ instead of $O(n^2)$.

\textbf{Average proximity vector:} The average proximity vector is a row vector of length $n$ in the average proximity matrix for each of the $n$ users. We maintain only a sparse approximation vector for each user, thus, only the non-zeroes entries are kept in memory as a set of pairs of type $(index,value)$. We define the proximity between the current location vectors $u,v\in\REAL^3$ of two users as: $prox(u,v):=e^{-\dist(u,v)}$.

\textbf{Coreset for a streamed average:} Whenever a new record $(time, user ID, longitude, latitude)$ is inserted to the stream, we update the entries for that user as his/her current position array $\pos$ is changed. Next, we compute that $n$ proximities from that user to each of the other users. Note that in our case, the proximity from a user to himself is always $e^0=1$. Each proximity $prox_{j}$ where $1\leq j\leq n$ is converted to a sparse vector $(0,\cdots,0,prox_j,0,\cdots,0)$ with one non-zero entry. This vector should be added to the average proximity vector of user $j$, to update the $j$th entry. Since maintaining the exact average proximity vector will take $O(n)$ memory for each of the $n$ users, we instead add this sparse vector to an object (``coreset") that maintains an approximation to the average proximity vector of user $j$.

Our problem is then reduced to the problem of maintaining a sparse average of a stream of sparse vectors in $\REAL^n$, using $O(\log n)$ memory. Maintaining such a stream for each of the $n$ users seen so far, will take overall memory of $O(n\log n)$ as desired, compared to the exact solution that requires $O(n^2)$ memory. We generalize and formalize this problem, as well as the approximation error and its practical meaning, in Section~\ref{sec:statement}.

\subsection{Related Work}
\label{sec:related-work}


As mobile applications become location-aware, the representation and analysis of location-based data sets become more important and useful in various domains \cite{wasserman1980analyzing, hogan2008analyzing, carrington2005models, dinh2010approximation, nguyen2011adaptive, lancichinetti2009community}. An interesting application is to extract the relationships between mobile users (in other words, their social network) from their location data \cite{liao2006location,zheng2011location, dinh2013adaptive}. Therefore, in this paper, we use coresets to represent and approximate (streaming) GPS-based location data for the extraction of the social graphs. The problem in social network extraction from GPS data is closely related to the frequency moment problem. {\it Frequency approximation} is considered the main motivation for streaming in the seminal work of~\cite{AMS}, known as the ``AMS paper", which  introduced the streaming model.

Coresets have been used in many related applications. The most relevant are coresets for $k$-means; see~\cite{barger2015k} and reference therein. Our result is related to coreset for $1$-mean that approximates the mean of a set of points. A coreset as defined in this paper can be easily obtained by uniform sampling of $O(\log d\log(1/\delta))/\eps^2$ or $O(1/(\delta\eps))$ points from the input, where $\delta\in(0,1)$ is the probability of failure. However, for sufficiently large stream the probability of failure during the stream approaches $1$. In addition, we assume that $d$ may be arbitrarily large. In~\cite{barger2015k} such a coreset was suggested but its size is exponential in $1/\eps$. We aim for a deterministic construction of size independent of $d$ and linear in $1/\eps$.

A special case of such coreset for the case that the mean is the origin (zero) was suggested in~\cite{mikhail}, based on Frank-Wolfe, with applications to coresets for PCA/SVD. In this paper we show that the generalization to any center is not trivial and requires a non-trivial embedding of the input points to a higher dimensional space.

Each of the above-mentioned prior techniques has at least one of the following disadvantages: (1) It holds only for positive entries. Our algorithm supports any real vector. Negative values may be used for deletion or decreasing of counters, and fraction may represent weights. (2) It is randomized, and thus will always fail on unbounded stream. Our algorithm is deterministic. (3) It supports only $s=1$ non-zero entries. Our algorithm supports arbitrary number of non-zeroes entries with only linear dependency of the required memory on $s$. (4) It projects the input vectors on a random subspace, which diminishes the sparsity of these vectors. Our algorithm maintains a small weighted subset of the vectors. This subset keeps the sparsity of the input and thus saves memory, but also allows us to learn the representative indices of points (time stamps in our systems) that are most important in this sense.

The most important difference and the main contribution of our paper is the error guarantee. Our error function in~\eqref{err} is similar to~$\norm{\bar{p}-\hat{p}}_{\ell}\leq \eps \norm{\bar{p}}_q$ for $\ell=2$ on the left hand side. Nevertheless, the error on the right hand side might be significantly smaller: instead of taking the sum of squared distances to the origin (norm of the average vector), we use the variance, which is the sum of squared distances to the mean. The later one is always smaller, since the mean of a set of vectors minimized their sum of squared distances.

\vspace{-0.3cm}
\section{Problem Statement}
\label{sec:statement}

The input is an unbounded stream vectors $p_1,p_2,\cdots$ in $\REAL^d$. Here, we assume that each vector has one non-zero entry. In the social network example, $d$ is the number of users and each vector is in the form $(0,\cdots,0,\prox_j,0,\cdots,0)$, where $prox_j$ is the proximity between the selected user and user $j$. Note that for each user, we independently maintain an input stream of proximities to each of the other users, and the approximation of its average. In addition, we get another input: the error parameter $N$, which is related to the memory used by the system. Roughly, the required memory for an input stream will be $O(N\log n)$ and the approximation error will be $\eps:=1/N$. That is, our algorithms are efficient when $N$ is a constant that is much smaller than the number of vectors that were read from stream, $2<N\ll n$. For example, to get roughly $1$ percents of error, we have $\eps=0.01$, and the memory is about $100n$, compared to $n^2$ for the exact average proximity.

The output $\hat{p}$ is an $N$-sparse approximation to the average vector in the stream over the $n$ vectors $p_1,\cdots,p_n$ seen so far.  That is, an approximation to the centroid, or center of mass, $\bar{p}=\frac{1}{n}\sum_i p_i$. Here and in what follows, the sum is over $i\in \br{1,\cdots,n}$. Note that even if $s=1$, the average $\bar{p}$ might have $n\gg N$ non-zero entries, as in the case where $p_i=(0,\cdots,0,1,0,\cdots,0)$ is the $i$th row of the identity matrix.
The sparse approximation $\hat{p}$ of the average vector $\bar{p}$ has the following properties: (1) The vector $\hat{p}$ has at most $N$ non-zero entries. (2) The vector $\hat{p}$ approximates the vector of average proximities $\bar{p}$ in the sense that the (Euclidean) distance between the two vectors is $\var/N$ where $\var$ is the variance of all the vectors seen so far in the stream.
    More formally, $\norm{\bar{p}-\hat{p}}_2 \leq \eps \var,$
    where $\eps=1/N$ is the error, $\bar{p}=\frac{1}{n}\sum_{i=1}^n p_i$ is the average vector in $\REAL^d$ for the $n$ input vectors, and the variance is the sum of squared distances to the average vector.

\textbf{Distributed and parallel computation. } Our system supports distributed and streaming input simultaneously in a  ``embarrassingly parallel" fashion. E.g., this allows multiple users to send their streaming smart-phone data to the cloud simultaneously in real-time. There is no assumption regarding the order of the data in user ID. Using $M$ nodes, each node will have to use only $1/M$ fraction of the memory to $(\log n)^{O(1)}/M$ that is used by one node for the same problem, and the average insertion time for a new point will be reduced by a factor of $M$ to $(\log n)^{O(1)}/M$.

Parallel coreset computation of unbounded streams of distributed data was suggested in~\cite{FT15}, as an extension to the classic merge-and-reduce framework in~\cite{sax,har2006coresets}. We apply this framework on our off-line algorithm to handle streaming and distributed data (see Section~\ref{sec:algorithms}).

\textbf{Generalizations. } Above we assume that each vector in the stream has a single non-zero entry. To generalize that, we now assume that each vector has at most $s$ non-zeroes entries and that these vectors are weighted (e.g. by their importance). Under these assumptions, we wish to approximate the weighted mean $\bar{p}=\sum_{i=1}^n u_i p_i$ where $u=(u_1,\cdots,u_n)$ is a weight vector that represents distribution, i.e., $u_i\geq 0$ for every $i\in[n]$. Then,  our problem is formalized as follows.
%
\begin{problem}
Consider an unbounded stream of real vectors $p_1,p_2,\cdots$, where each vector is represented only by its non-zero entries, i.e., pairs $(entryIndex, value)\in \br{1,2,3,\ldots}\times \REAL$. Maintain a subset of $N\ll n$ input vectors, and a corresponding vector of positive reals (weights) $w_1,w_2,\cdots, w_{N}$, where the sum $\hat{p}:=\sum_{i=1}^N w_ip_i$ approximates the sum  $\bar{p}=\sum_{i=1}^n p_i$ of the $n$ vectors seen so far in the stream up to a provably small error that depends on its variance $\var(p):=\sum_{i=1}^n \norm{p_i-\bar{p}}_2^2$.
Formally, for an error parameter $\eps$ that may depend on $N$,
\begin{equation}\label{err}
\norm{\bar{p}-\hat{p}}\leq \eps\var(p).
\end{equation}
\end{problem}
\vspace{-0.3cm}
We provide a solution for  this problem mainly by proving Theorem~\ref{thm:coreset} for off-line data, and turn it into algorithms for streaming and distributed data as explained in Section~\ref{sec:algorithms}.

 \begin{figure}[t!]
\centering
\includegraphics[width=.97\linewidth, height=3cm]{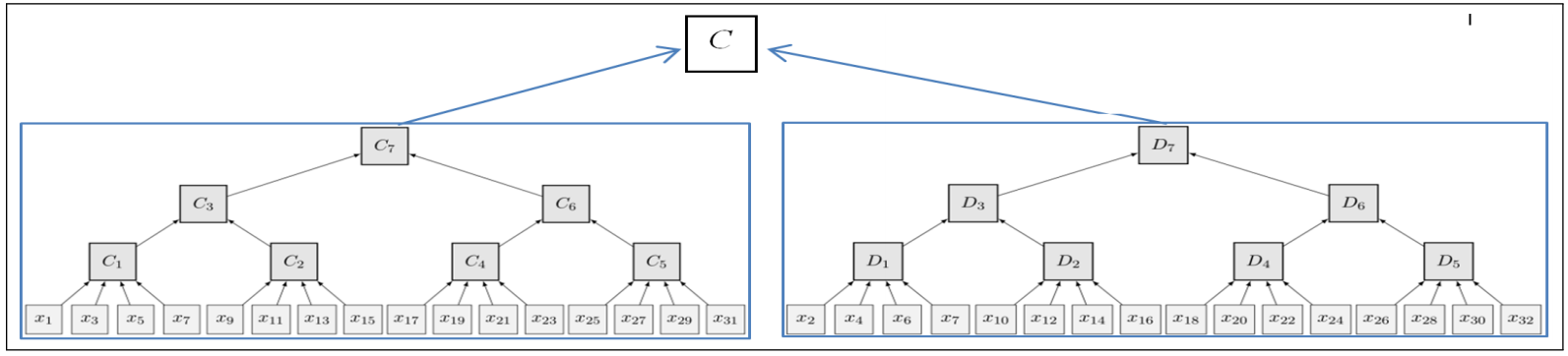}
\vspace{-0.2in}
\caption{Coreset computation of streaming data that is distributed among $M=2$ machines.
The odd/even vectors in the stream (leaves) are compressed by the machine on the left/right, respectively.
A server (possibly one of these machine) collects the coreset $C_7$ and $D_7$ from each machine to obtain the final coreset $C$ of the $n=32$ vectors seen so far. Each level of each tree stores at most one coreset in memory, and overall of $O(\log n)$ coresets.}
 \vskip -6pt
\label{fig:tree}
\end{figure}

\section{New Coreset Algorithms}
\label{sec:algorithms}

In this section we first describe an algorithm for approximating the sum of $n$ streaming vectors using one pass. The algorithm calls its off-line version as a sub-procedure. We then explain how to run the algorithm on distributed \emph{and} unbounded streaming data using $M$ machines or parallel threads. The size of the weighted subset of vectors that are maintained in memory, and the insertion time per new vector in the stream are logarithmic on the number $n$ of vectors in the stream. Using $M$ machines, the memory and running time per machine is reduced by a factor of $M$.

\subsection{Streaming and Distributed Data}
\textbf{Overview.} The input to Algorithm 1 is a $\stream$ provided as a pointer to a device that sends the next input vectors in a stream that consists of $n$ vectors, upon request, For example, a hard drive, a communication socket, or a web-service that collects information from the users. The second parameter $\eps$ defines the approximation error. The required memory grows linearly with respect to $1/\eps$.

Algorithm 1 maintains a binary tree whose leaves are the input vectors, and each inner node is a coreset, as in the left or right hand side of Fig.~\ref{fig:tree}. However, at most one coreset in each level of the tree is actually stored in memory. In Line~\ref{a1l1}, we initialize the current height of this tree. Using $\log(n)/\eps$ vectors in memory our algorithm returns an $O(\eps)$-coreset, but to get exactly $\eps$-coreset, we increase it in Line 2 by a constant factor $\alpha$ that can be find in the proof of Theorem~\ref{thm:coreset}. In Lines 3-14, we read batches (sets) of $O(\log(n)/\eps)$ vectors from the stream and compress them. The last batch may be smaller. Line 4 defines the next batch $P$. Unlike the coresets in the nodes of the tree, we assume that the input vectors are unweighted, so,  Line 5 defines a weight $1$ for each input vector. Line 6 reduce the set $P$ by half to the weighted coreset $(S,w)$ using Algorithm 2 (the off-line coreset construction.) Theorem~\ref{thm:coreset} guarantees that such a compression is possible.

In Lines~\ref{a1l8}--\ref{a1l12} we add the new coreset $(S,w)$ to the lowest level $\ell$ of the binary tree, if it is not assigned to a coreset already, i.e., $S_{\ell}$ is not empty. Otherwise, we merge the new coreset $(S,w)$ with the level's coreset $S_{\ell}$, mark the level as empty from coresets ($S_{\ell}\gets \emptyset$), and continue to the next higher level of the tree until we reach a level $\ell$ that is not assigned to a coreset , i.e., $S_{\ell}=\emptyset$. Line~\ref{a1l13} handle the case where we reach to the root of the tree, and a new (top) level is created. In this case, only the new root of the tree contains a coreset.

When the streaming is over, in Lines~\ref{a1l15}--\ref{a1l16} we collect the active coreset in each of the $O(\log n)$ tree levels (if it has one) and return the union of these coresets.

\textbf{Parallel computation on distributed data.} In this model each machine has its own stream of data, and computes its coreset independently and in parallel to the other machines, as described above. Whenever we wish to get the coreset for the union of streaming vectors, we collect the current coreset from each machine on a single machine or a server. Since each machine sends only a coreset, the communication is also logarithmic in $n$. For $M\geq 2$ machines and a single input stream, we send every $i$th point in the stream to the $i$th machine, for every $i$ between $1$ and $M$. For example, if $M=2$, the odd vectors will be sent to the first machine, and every second (even) vector will be sent to the second machine. Then, each machine will compute a coreset for its own unbounded stream; See Fig~\ref{fig:tree}.

\subsection{Off-line data reduction}
\label{sec:offline-algorithm}

Algorithm 2 is called by Algorithm 1 and its variant in the last sub-section for compressing a given small set $P$ of vectors in memory by computing its coreset. As in Line~\ref{a2l10} of Algorithm 1, the input itself might be a coreset of another set, thus we assume that the input has a corresponding weight vector $u$. Otherwise, we assign a weight $1$ for each input vector, i.e., $u=(1,\cdots,1)$. The output of Algorithm 1 is an $\eps$-coreset $(S,w)$ of size $|S|=O(1/\eps)$ for a given $\eps\in(0,1)$. While the number $n$ of input vectors can be of an arbitrary size, Algorithm 2 always passes an input set of $n=2|S|$ points to get output that is smaller by half.

\textbf{Overview of Algorithm 2:} In Line~\ref{a2l1} the desired mean $E_u$ that we wish to approximate is computed.
Lines~\ref{a2l2}--\ref{a2l4} are used for adding an extra dimension for each input vector later. In Lines 4-6 we normalize the augmented input, by constructing a set $q_1,\cdots,q_n$ of unit vectors with a new set of weights $s_1,\cdots,s_n$ whose mean is $\sum_i s_i q_i=(0,\cdots,0,x/v)$. We then translate this mean to the origin by defining the new set $H$ in Line~\ref{a2l7}.

The main coreset construction is computed in Lines~\ref{a2l9}--\ref{a2l13} on the normalized set $H$ whose mean is the origin and its vectors are on the unit ball. This is a greedy, gradient descent method, based on the Frank-Wolfe framework~\cite{mikhail}. In iteration $i=1$, we begin with an arbitrary input point $c_1$ in $H$. Since $c_1$ is a unit vector, its distance from the mean of $H$ (origin) is $1$. In Line~\ref{a1l11}--~\ref{a1l12} we compute the farthest point $h_2$ from $c_1$, and approximates the mean using only $c_1$ and the new point $h_2$. This is done, by projecting the origin on the line segment through $c_1$ and $h_2$, to get an improved approximation $c_2$ that is closer to the origin. We continue to add input points in this manner, where in the $i$th iteration another input point is selected for the coreset, and the new center is a convex combination of $i$ points. In the proof of Theorem~\ref{thm:coreset} it is shown that the distance to the origin in the $i$th iteration is $\alpha/i$, which yields an $\eps$-approximation after $\beta=O(\alpha/\eps)$ iterations.

The resulting center $c_\beta$ is spanned by $\beta$ input vectors. In Line~\ref{a2l11} we compute their new weights based on their distances from $c_\beta$. Lines~\ref{a2l14}--\ref{a2l16} are used to convert the weights of the vectors in the normalized $H$ back to the original input set of vectors. The algorithm then returns the small subset of $\beta$ input vectors with their weights vector $w$.

\newcommand{\scoreset}{\textsc{Streaming-Coreset}}
\newcommand{\coresetsize}{coresetSize}
\begin{algorithm2e}[h]
\label{a1}
    \caption{$\scoreset(\stream, \eps)$}
\begin{tabbing}
\textbf{Input:} \quad\quad\=An input stream of $n$ vectors in $\REAL^d$.\\
\> an error parameter $\eps\in(0,1)$\\
\textbf{Output:} \>An $\eps$-coreset $(S,w)$ for the set of $n$ vectors; \\
\> see Theorem 1.
\end{tabbing}
\vspace{-0.3cm}
\nl Set $max\gets 0$\label{a1l1}\\
\nl Set $\alpha$ to be a sufficiently large constant that can be derived from the proof of Theorem 1.  \label{a1l2}\\
\nl \While{$\stream$ \textit{is not empty}\label{a1l3}}{
	\nl Set $P\gets$ next $\lceil 2\alpha\ln(n)/\eps\rceil$ input vectors in $\stream$ \label{a1l4}\\
    \nl Set $u\gets (1,\cdots,1)$ where $u$ has $|P|$ entries. \label{a1l5}\\
	\nl Set $(S,w) \gets \coreset(P, u,\eps/(\alpha \ln(n))$\label{a1l6} \\
	\nl Set $\ell\gets 1$ \label{a1l7}\\
	\nl \While{$S_\ell\neq \emptyset$ and $\ell\leq max$ \label{a1l8}} {
	\nl Set $S_\ell \gets S \cup S_\ell\label{all7}$ \label{a1l9}\\
	\nl Set $(S,w) \gets \coreset(S_{\ell}, w_\ell, \eps)$ \label{a1l10}\\
	\nl Set $S_\ell\gets \emptyset$\label{a1l11}\\
	\nl Set $\ell\gets \ell+1$\label{a1l12}
	}
    \nl \If{$\ell>max$}{Set $max\gets \ell$}\label{a1l13}
	\nl Set $(S_\ell ,w_\ell)\gets (S,w)$\label{a1l14}
}
    \nl Set $S\gets\bigcup_{i=1}^{max} S_i$ and $w\gets (w_1,w_2,\cdots,w_{max})$\label{a1l15}\\
	\nl \Return {$(S,w)$}\label{a1l16}\\
\end{algorithm2e}

\begin{algorithm2e}[t!]
\label{a2}
    \caption{$\coreset(P, u,\eps)$}
\begin{tabbing}
\textbf{Input:} \quad\quad\=A set $P$ of vectors in $\REAL^d$, \\
\> a positive weight vector $u=(u_1,\cdots,u_n)$, \\
\> an error parameter $\eps\in(0,1)$\\
\textbf{Output:} \>An $\eps$-coreset $(S,w)$ for $(P,u)$
\end{tabbing}
\vspace{-0.3cm}
\nl Set $E_u\gets \sum_{i=1}^n u_ip_i$ \label{a2l1}\\
\nl Set $x\gets \sum_{j=1}^n u_j \norm{p_j-E_u}$ \label{a2l2}\\
\nl Set $v\gets\sum_{j=1}^n u_j \norm{(p_j-E_u,x)}$\label{a2l3}\\
\nl \For{$i\gets 1$ \textbf{\emph{to}} $n$\label{a2l4}}{
\nl Set $\displaystyle q_i\gets \frac{(p_i-E_u,x)}{\norm{(p_i-E_u,x)}}$\label{a2l5}\\
\nl Set $\displaystyle s_i\gets \frac{u_i\norm{(p_i-E_u,x)}}{v}\label{a2l6}$
}
\nl Set $H\gets \br{q_i-(0,\cdots,0,x/v)\mid i\in[n]}$\label{a2l7}\\
\nl Set $\alpha\gets$ a sufficiently large constant that can be derived from the proof of Theorem 1.  \\
\nl Set $c_1\gets$ an arbitrary vector in $H$\label{a2l9}\\
\nl \For{$i\gets 1$ to $\beta:=\lceil \alpha/\eps\rceil$\label{a2l10}}{
\nl   $h_{i+1}\gets$ farthest point from $c_i$ in $H$\label{a2l11}\\
\nl    $c_{i+1}\gets$ the projection of the origin on the segment $\overline{c_i, h_{i+1}}$\label{a2l12}
}
\nl Compute $w'=(w'_1,\cdots,w'_{\beta})\in S^\beta$ such that $c_{\beta}=\sum_{i=1}^{\beta} w'_i h_{i+1}$\label{a2l13}\\
\nl \For{$i\gets 1$ to $\beta$\label{a2l14}}{
\nl $\displaystyle w''i\gets \frac{vw'_i}{(p_i-E_u,x)}$\label{a2l15}\\
\nl $\displaystyle  w_i\gets \frac{w''_i}{\sum_{j=1}^\beta w''_j}$\label{a2l16}\\
}
\nl $w\gets (w_1,\cdots,w_\beta)$ \label{a2l17}\\
\nl $S\gets \br{v_1,\cdots,w_\beta}$\label{a2l18}\\
\nl \Return $(S,w)$\label{a2l19}
\end{algorithm2e}

\vspace{-0.4cm}
\section{Correctness}
\label{sec:correctness}

In this section we show that Algorithm 1 computes correctly the coreset for the average vector in Theorem~\ref{thm:coreset}. 

Let $D^n$ denote all the possible distributions over $n$ items, i.e., $D$ is the unit simplex
\[\Scale[0.9]{D^n=\br{(u_1,\cdots,u_n)\in\REAL^n \mid u_i\geq 0 \text{ and } \sum_{i=1}^n u_i=1}.}
\]

Given a set $P=\br{p_1,\cdots,p_n}$ of vectors, the mean of $P$ is $\frac{1}{n}\sum_{i=1} p_i$. This is also the expectation of a random vector that is chosen uniformly at random from $P$.  The sum of variances of this vector is the sum of squared distances to the mean.
More generally, for a distribution $u\in S$ over the vectors of $P$, the (weighted) mean is $\sum_{i=1}^n u_ip_i$, which is the expected value of a vector chosen randomly using the distribution $u$. The variance $\var_u$ is the sum of weighted squared distances to the mean. By letting $N=1/\eps$ in the following theorem, we conclude that there is always a sparse distribution $w$ of at most $1/\eps$ non-zeroes entries, that yields an approximation to its weighted mean, up to an $\eps$-fraction of the variance.
\vspace{-0.3cm}
\begin{theorem}[Coreset for the average vector]
Let $u\in D^n$ be a distribution over a set $P=\br{p_1,\cdots,p_n}$ of $n$ vectors in $\REAL^d$, and let $N\geq 1$. Let $(S,w)$ denote the output of a call to $\coreset(P,u,1/N)$; see Algorithm 1. Then $w\in D^n$ consists $O(N)$ non-zero entries, such that the sum $\bar{p}=\sum_{i=1} u_i p_i$ deviates from the sum $\hat{p}= \sum_{i=1}^n w_i p_i$ by at most a $(1/N)$-fraction of the variance $\var_u=\sum_{i=1}^n u_i\norm{p_i-\bar{p}}_2^2$, i.e., ${\norm{\bar{p}-\hat{p}}_2^2 \leq \frac{\var_u}{N}}$.
\label{thm:coreset}
\end{theorem}
By the $O(\cdot)$ notation above, it suffices to prove that there is a constant $\alpha>0$ such that $N\geq \alpha$ and
\begin{equation}\label{eq0}
\Scale[0.9]{\norm{E_u-E_w}_2^2 \leq \frac{\alpha\var_u}{N},}
\end{equation}
where $E_u=\bar{p}$ and $E_w=\hat{p}$. The proof is constructive and thus immediately implies Algorithm 1. Indeed, let \[
\Scale[0.9]{x=\sum_j u_j \norm{p_j-E_u}, \text{ and } v=\sum_j u_j \norm{(p_j-E_u,x)}}.
\]
Here and in what follows, $\norm{\cdot}=\norm{\cdot}_2$ and all the sums are over $[n]=\br{1,\cdots,n}$.
For every $i\in [n]$ let
\[
\Scale[0.9]{q_i=\frac{(p_i-E_u,x)}{\norm{(p_i-E_u,x)}}}, \text{ and } \Scale[0.9]{s_i=\frac{u_i\norm{(p_i-E_u,x)}}{v}}.
\]
Hence,
\vspace{-0.3cm}
\begin{equation}\label{eq1}
\Scale[0.8]{
\begin{aligned}
\sum_i s_i q_i&=\frac{1}{v} \sum_i u_i (p_i -E_u,x)\\
&= \frac{1}{v} \left(\sum_i u_ip_i-\sum_m u_mE_u,\sum_k u_kx  \right)\\
&=\frac{1}{v}\left(\sum_i u_i p_i -\sum_j u_jp_j,x\right)=\left(0,\cdots,0,\frac{x}{v} \right).
\end{aligned}}
\end{equation}
\normalsize
Since $(s_1,\cdots,s_n)\in D^n$ we have that the point $p=\sum_i s_i q_i$ is in the convex hull of $Q=\br{q_1,\cdots,q_n}$.
By applying the Frank-Wolfe algorithm as described in~\cite{clarkson} for the function $f(s)=\norm{As}$, where each row of $A$ corresponds to a vector in $Q$, we conclude that there is $w'=(w'_1,\cdots,w'_n)\in D^n$ that has at most $N$ non-zero entries such that
\vspace{-0.3cm}
\begin{equation}\label{eq2}
\Scale[0.8]{
\norm{\sum_i (s_i-w'_i)q_i}^2=\norm{\sum_i s_i q_i -\sum_j w'_jq_j}^2=\norm{p-q}^2\leq \frac{1}{N}.}
\end{equation}
\vspace{-0.2cm}
For every $i\in[n]$, define
\[
w''_i=\frac{vw'_i}{\norm{(p_i-E_u,x)}}
\text{\quad and \quad}w_i = \frac{w''_i}{\sum_j w''_j}.
\]
\vspace{-0.2cm}
We thus have:
\vspace{-0.2cm}
\small

\begin{equation}
\label{eq33}
\Scale[0.9]{
 \norm{E_u-E_w}^2=\norm{\sum_i u_ip_i -\sum_j w_jp_j}^2=\norm{\sum_i (u_i-w_i)p_i}^2}
\end{equation}
\begin{equation}
\label{eq4}
\Scale[0.9]{
=\norm{\sum_i (u_i-w_i)(p_i-E_u,x)}^2}
\end{equation}
\begin{equation}\Scale[0.9]{
=v^2\norm{\sum_i \left(\frac{u_i\norm{(p_i-E_u,x)}}{v}-\frac{w_i\norm{(p_i-E_u,x)}}{v} \right)q_i}^2}
\end{equation}
\begin{equation}
\label{eq55}
\Scale[0.9]{
=v^2\norm{\sum_i \left(s_i-\frac{(w_i''/\sum_j w''_j)\cdot \norm{(p_i-E_u,x)}}{v}\right)q_i}^2}
\end{equation}
\begin{equation}
\label{eq6}
\Scale[0.9]{
=v^2\norm{\sum_i \left(s_i-\frac{w'_i}{\sum_j w''_j} \right)q_i}^2,}
\end{equation}


\normalsize
where \eqref{eq33} is by the definitions of $E_u$ and $E_w$,~\eqref{eq4} follows since $\sum_i u_i=\sum_i w_i=1$ and thus $\sum_i u_iy=\sum_j u_jy$ for every vector $y$,~\eqref{eq55} follows by the definitions of $w_i$ and $q_i$, and~\eqref{eq6} by the definition of $w'_i$. Next, we bound~\eqref{eq6}. Since for every two reals $y,z$,
\begin{equation}\label{eqs}
2yz\leq y^2+z^2
\end{equation}
 by letting $y=\norm{a}$ and $z=\norm{b}$ for $a,b\in\REAL^d$,
 \small
\begin{equation}\label{eq7}
\Scale[0.85]{
\begin{aligned}
\norm{a+b}^2&\leq \norm{a}^2+\norm{b}^2+2\norm{a}\norm{b}\\
&\leq\norm{a}^2+\norm{b}^2+(\norm{a}^2+\norm{b}^2)=2\norm{a}^2+2\norm{b}^2.
\end{aligned}
}
\end{equation}
\normalsize
By substituting $a=\sum_i (s_i-w'_i)q_i$ and $b=\sum_i (w'_i-\frac{w'_i}{\sum_j w''_j})q_i$ in~\eqref{eq7}, we obtain
\small
\begin{align}
\norm{\sum_i \left(s_i-\frac{w'_i}{\sum_j w''_j}\right)q_i}^2
\label{eq8}&\leq 2\norm{\sum_i (s_i-w'_i)q_i}^2\\
\label{eq9}&\quad+2\norm{\sum_i \left(w'_i-\frac{w'_i}{\sum_j w''_j} \right)q_i}^2.
\end{align}
\normalsize
\vspace{-0.2in}
\paragraph{Bound on~\eqref{eq9}:} Observe that
\small
\begin{align}
\nonumber\norm{\sum_i \left(w'_i-\frac{w'_i}{\sum_j w''_j} \right)q_i}^2
&=\norm{\sum_i w'_i\left(1-\frac{1}{\sum_j w''_j}\right)}^2\\
\label{eq10}&=\left(1-\frac{1}{\sum_j w''_j}\right)^2\cdot \norm{\sum_i w'_i q_i}^2.
\end{align}
\normalsize
Let $\tau=\frac{v}{\sqrt{N}x}$. By the triangle inequality
\begin{equation}\label{eq11}
\Scale[0.85]{v=\sum_j u_j\norm{(p_j-E_u,x)}\leq \sum_j u_j \norm{p_j-E_u}+x=2x.}
\end{equation}
By choosing $c>16$ in~\eqref{eq0} we have $N\geq 16$, so
\begin{equation}\label{eq12}
\Scale[0.9]{\tau \leq \frac{2}{\sqrt{N}}\leq \frac{1}{2}.}
\end{equation}
Substituting $a=-\sum_i s_i q_i$ and $b=\sum_i (s_i-w'_j)q_j$ in~\eqref{eq7} bounds the right expression of~\eqref{eq10} by
\small
\begin{equation}\label{eq13}
\Scale[0.85]{
\begin{aligned}
\norm{\sum_i w'_iq_i}^2 &\leq 2\norm{\sum_i s_iq_i}^2 + 2\norm{\sum_i (s_i-w'_j)q_j}^2\\
&\leq \frac{2x^2}{v^2}+\frac{2}{N}=\frac{2(1+\tau^2)}{\tau^2N},
\end{aligned}
}
\end{equation}
\normalsize
where the last inequality follows from~\eqref{eq1} and~\eqref{eq2}. For bounding the left expression of~\eqref{eq10}, note that
\small
\begin{align}
\label{eq14} (1-\sum_j w''_j)^2&=\left(1-\sum_j w'_j \cdot\frac{v}{\norm{(p_j-E_u,x)}} \right)^2\\
\nonumber &\leq \norm{(0,\cdots,0,1)-\sum_j w'_j\cdot \frac{(\frac{v}{x}(p_j-E_u),v)}{\norm{(p_j-E_u,x)}}}^2 \\
\nonumber &=\frac{v^2}{x^2} \norm{\left(0,\cdots,0,\frac{x}{v}\right)-\sum_j w'_j \cdot \frac{(p_j-E_u,x)}{\norm{(p_j-E_u,x)}}}^2\\
\nonumber &= \frac{v^2}{x^2}\norm{\sum_i (s_i-w'_i)q_i}^2\leq \frac{v^2}{Nx^2}=\tau^2,
\end{align}
\normalsize
where~\eqref{eq14} follows since $\norm{b}^2\leq \norm{(a,b)}^2$ for every pair $a,b$ of vectors, and the last inequality is by~\eqref{eq1} and~\eqref{eq2}. Hence, $\sum_j w''_j \geq 1-\tau$, and
\small
\[\Scale[0.9]{
\left(1-\frac{1}{\sum_j w''_j}\right)^2
=\left(\frac{1-\sum_i w''_i}{\sum_j w''_j}\right)^2\leq \frac{\tau^2}{(1-\tau)^2}.}
\]
\normalsize
Combining~\eqref{eq10} and~\eqref{eq13} bounds~\eqref{eq9} by
\small
\[\Scale[0.85]{
\begin{split}
2\norm{\sum_i \left(w'_i -\frac{w'_i}{\sum_j w''_j}\right)q_i}^2
=2\left(1-\frac{1}{\sum_j w''_j}\right)^2\cdot \norm{\sum_i w'_i q_i}^2\\
\leq \frac{2\tau^2}{(1-\tau)^2}\cdot \frac{2(1+\tau^2)}{\tau^2N}=\frac{4(1+\tau^2)}{N(1-\tau)^2}.
\end{split}}
\]
\normalsize

 \begin{figure}[t!]
\centering
\includegraphics[width=.8\linewidth]{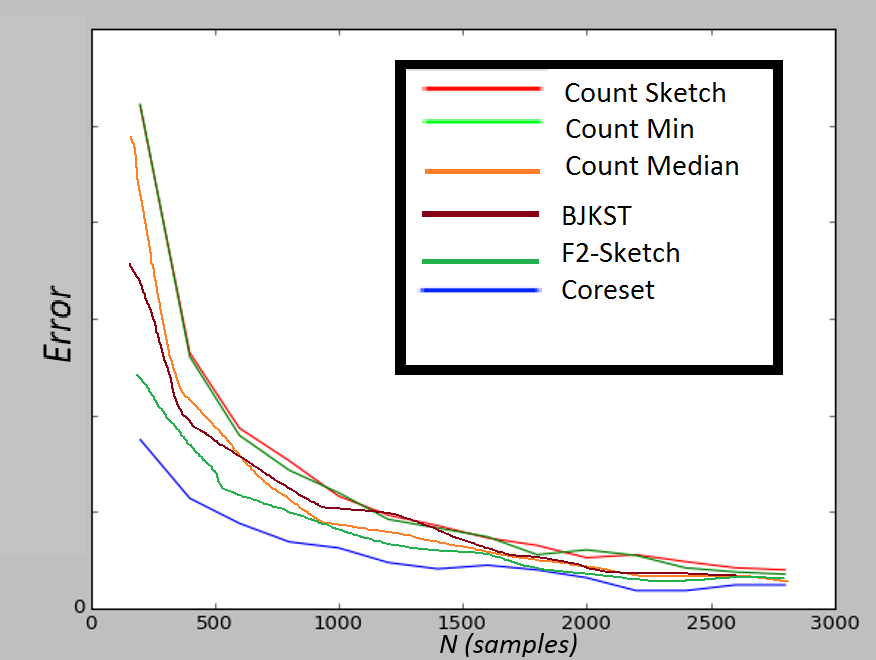}
\vspace{-0.19in}
\caption{Given a set of vectors from a standard Gaussian distribution, the graph shows the $\ell_2$ error (y-axis) between their sum and their approximated sum using only $N$ samples (the $x$-axis) based on Count Sketch~\cite{ccfc04}, Count Min~\cite{Cm05}, Count Median~\cite{cormode2005improved}, BJKST~\cite{bar2002counting}, F2-Sketch~\cite{alon1996space}, and our coreset.}
 \vskip -6pt
\label{fig:small}
\end{figure}

\vspace{-0.18in}

\paragraph{Bound on~\eqref{eq8}:} Plugging the last inequality,~\eqref{eq2} and~\eqref{eq8} in~\eqref{eq6} yields
\small
\vspace{-0.2in}
\begin{equation}\label{eq15}
\Scale[0.85]{
\begin{split}
\norm{E_u-E_w}^2&=v^2\norm{\sum_i \left(s_i-\frac{w'_i}{\sum_j w''_j} \right)q_i}^2\\
&\leq 2v^2 \Big(\norm{\sum_i (s_i-w'_i)q_i}^2 \quad+\norm{\sum_i \left(w'_i-\frac{w'_i}{\sum_j w_j''}\right)q_i}^2\Big)\\
&\leq \frac{2v^2}{N}\left(1+\frac{2(1+\tau^2)}{(1-\tau)^2} \right)\leq \frac{\alpha v^2}{N},
\end{split}
}
\end{equation}
\normalsize
for a sufficiently large constant $\alpha$, e.g. $\alpha=3$, where in the last inequality we used~\eqref{eq12}. Since $v\leq 2x$ by~\eqref{eq11} we have

\small
\[\Scale[0.85]{
\begin{split}
v&\leq 2x=2\sum_j u_j \norm{p_j-E_u}\\
&=\sum_j 2\cdot \sqrt{\sqrt{\var_u}}\sqrt{u_j}\cdot\frac{\sqrt{u_j}\norm{p_j-E_u}}{\sqrt{\sqrt{\var_u}}}\\
&\leq \sum_j \left(\sqrt{\var_u}u_j +\frac{u_j\norm{p_j-E_u}^2}{\sqrt{\var_u}}\right)\\
&=\sqrt{\var_u}+\frac{1}{\sqrt{\var_u}}\sum_j u_j \norm{p_j-E_u}^2=2\sqrt{\var_u},
\end{split}
}
\]
\normalsize
where in the second inequality we used~\eqref{eqs}.
Plugging this in~\eqref{eq15} and replacing $N$ by $4\alpha N=O(N)$ in the proof above, yields the desired bound
\small
\[\Scale[0.9]{
\norm{E_u-E_w}^2\leq \frac{\alpha v^2}{N}\leq \frac{4\alpha\cdot \var_u}{N}.
}
\]
\normalsize


 \begin{figure}
\centering
\includegraphics[width=.99\linewidth]{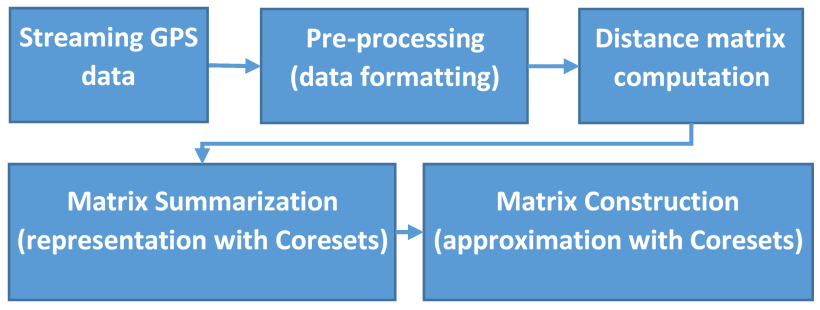}
\vspace{-0.2in}
\caption{The overview of our designed system to extract and represent social networks is given.}
\vskip -6pt
\label{fig:systemoverview}
\end{figure}



\vspace{-0.19in}
\section{Experimental Results}
\label{sec:experiments}
We implemented the coreset algorithms in 1 and 2. We also implemented a brute force method for determining the social network that considers the entire data. We used this method to derive the ground truth for the social network for small scale data. Our system's overview is given in Figure~\ref{fig:systemoverview} and explained in Section~\ref{sec:streamed-social-network}. The coreset algorithm computes the heavy hitters by approximating the sum of the columns of the proximity matrix as explained in Section~\ref{sec:streamed-social-network}.
 \begin{figure}
\centering
\includegraphics[width=.5\linewidth]{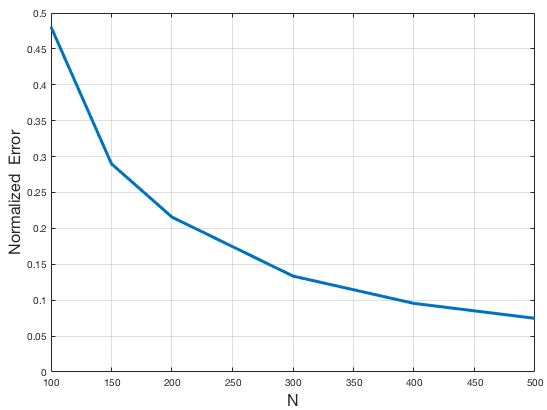}
\vspace{-0.2in}
\caption{The normalized error (y-axis) of proximities for the GPS traces taxi-drivers in the New York City Cab dataset using only $N$ samples (N increases along x-axis).}
\vskip -6pt
\label{fig:nyc}
\end{figure}
In this section, we have used different data sets from three different sources: In our first experiment, we compared our coreset algorithm's error with other sketch algorithms \cite{ccfc04, Cm05}. The second dataset is the New York City Cab dataset\footnote{https://publish.illinois.edu/dbwork/open-data/} and the third data set is from Stanford\footnote{https://snap.stanford.edu/data/} and includes six different graph-based data sets. In all the figures shown in this section, the x axis shows the coreset size (N) and the y axis represents the normalized error value $(Error*mean(var(p_i)^2)/mean(norm(p_i)))$, where $Error =  \norm{p_i-\bar{p}}_2$.
In our experiments, we ran N iterations and wrote down the empirical error $\epsilon$.

\textbf{Comparison to sketch algorithms:} Since the algorithms in  \cite{ccfc04, Cm05} focused on selecting the entries at scalar level (i.e., individual entries from a vector), in this experiment, we generated a small scale synthetic data (standard Gaussian distribution) and compared the error made by our coreset implementation to four other sketch implementations. These sketch algorithms are: Count Sketch, Count Min, Count Median, BJKST and F2-Sketch (see Fig.~\ref{fig:small}). For the sketch algorithms, we used the code available at\footnote{https://github.com/jiecchen/StreamLib/}. We plot the results in Fig.~\ref{fig:small} where our Coresets algorithm showed better approximation than all other well known sketch techniques for all N values.

\textbf{Application on NYC data:} Here we applied our algorithm on the NYC data. The data contains the location information of 13249 taxi cabs with 14776616 entries. The goal here is showing how the error on Coreset approximation would change on real data with respect to N (we expect that the error would reduce with respect to $N$ as the theory suggests). This can be seen in Fig.~\ref{fig:nyc}, where x axis is the coreset size (N) and y axis is the normalized error.

\begin{figure}
\begin{subfigure}{0.24\textwidth}
  \centering
  \includegraphics[width=0.92\linewidth]{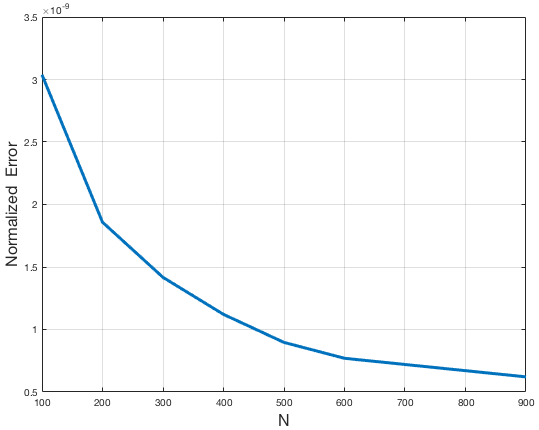}
    \vspace{-0.1in}
  \caption{Amazon data}
  \label{fig:sfig1}
\end{subfigure}%
\begin{subfigure}{0.24\textwidth}
  \centering
  \includegraphics[width=0.9\linewidth]{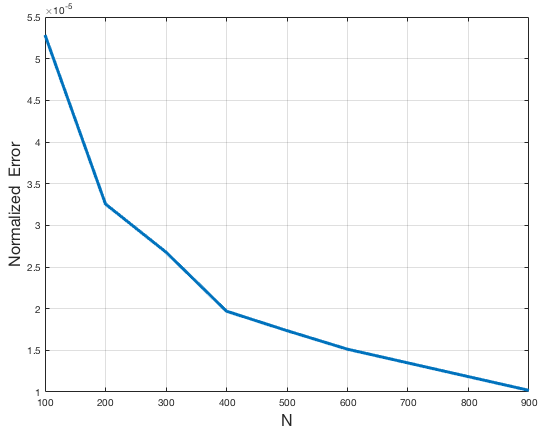}
  \vspace{-0.1in}
  \caption{YouTube data}
  \label{fig:sfig2}
\end{subfigure}

\begin{subfigure}{0.23\textwidth}
  \centering
  \includegraphics[width=0.92\linewidth]{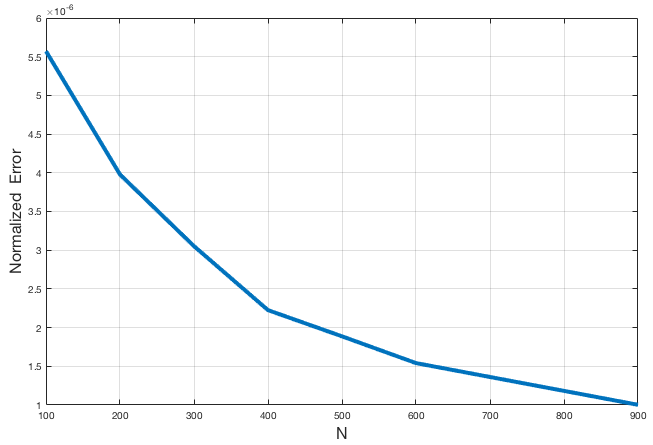}
    \vspace{-0.1in}
  \caption{DBLP data}
  \label{fig:sfig3}
\end{subfigure}
\hfill
\begin{subfigure}{0.23\textwidth}
  \centering
  \includegraphics[width=0.95\linewidth]{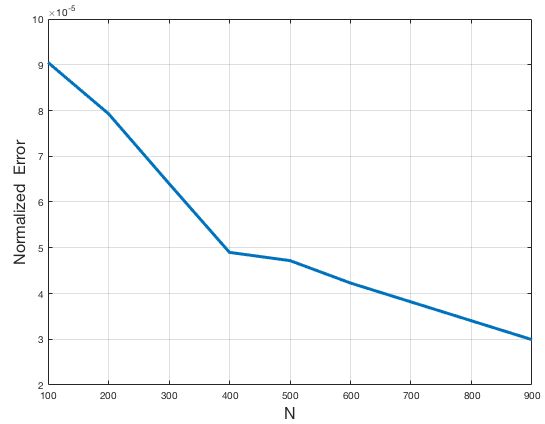}
    \vspace{-0.1in}
  \caption{Wikitalk data}
  \label{fig:sfig4}
\end{subfigure}

\begin{subfigure}{0.23\textwidth}
  \centering
  \includegraphics[width=0.92\linewidth]{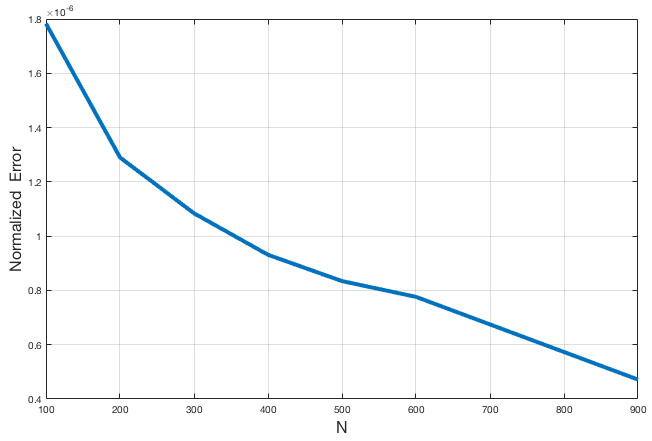}
    \vspace{-0.1in}
  \caption{Orkut data}
  \label{fig:sfig5}
\end{subfigure}
\hfill
\begin{subfigure}{0.23\textwidth}
  \centering
  \includegraphics[width=0.95\linewidth]{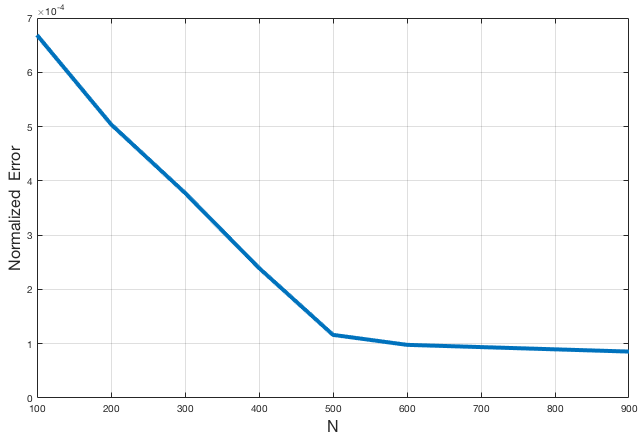}
    \vspace{-0.1in}
  \caption{LiveJournal data}
  \label{fig:sfig6}
\end{subfigure}

\vspace{-0.15in}
\caption{The normalized error (y-axis) of the coreset for network structure approximation is shown for four different datasets using only $N$ samples (N increases along x-axis).}
\label{fig:Stanford}
\end{figure}

\textbf{Application on Stanford Data Sets:} Here we apply our algorithm on six different data sets from Stanford: Amazon, Youtube,  DBLP, LiveJournal, Orkut and Wikitalk data sets. We run the Coreset algorithm to approximate the total number of connectivities each node has. We computed the error for each of the seven different N values from [100, 200, 300, 400, 500, 600, 900] for each data set. We used the first 50000 entries from the Orkut, Live Journal, Youtube and Wiki data sets and the first 5000 entries from Amazon and DBLP data set. The results are shown in Figure ~\ref{fig:Stanford}. In the figures, y axis represents the normalized error. The results demonstrate the utility of our proposed method for summarization.





\section{Conclusion}
In this paper we proposed a new coreset algorithm for streaming data sets with applications to summarizing large networks to identify the "heavy hitters". The algorithm takes a stream of vectors as input and maintains their sum using small memory. Our presented algorithm shows better performance at even lower values of non-zero entries (i.e., at higher sparsity rates) when compared to the other existing sketch techniques. We demonstrated that our algorithm can catch the heavy hitters efficiently in social networks from the GPS-based location data and in several graph data sets from the Stanford data repository.

\section*{Acknowledgements}

Support for this research has been provided in part by Ping An Insurance and NSFSaTC-BSF CNC 1526815. We are grateful for this support.
\bibliography{mybib1_1}
\bibliographystyle{icml2017}

\end{document}